\documentclass[10pt, conference, letterpaper]{IEEEtran}
\usepackage{cite}
\usepackage{amsmath,amssymb,amsfonts}
\usepackage{algorithmic}
\usepackage{booktabs,multirow}
\usepackage{graphicx}
\usepackage{textcomp}
\usepackage{xcolor}
\usepackage{tikz}
\usepackage{placeins}
\usetikzlibrary{positioning,arrows.meta}
\usetikzlibrary{arrows.meta,calc,positioning}
\usepackage{array}
\usepackage[ruled,vlined,linesnumbered]{algorithm2e}
\usepackage[hidelinks]{hyperref} 
\begin{document}

\title{Backdoors Leave Structural Traces: FedMAST for Backdoor Detection and Containment in Federated Learning}

\author{\IEEEauthorblockN{Srinivasan Subramanian}
\IEEEauthorblockA{\textit{Department of Computer Science} \\
\textit{Kennesaw State University}\\
Marietta, USA \\
ssubram7@students.kennesaw.edu}
\and
\IEEEauthorblockN{Md.~Abdullah~Al~Hafiz~Khan}
\IEEEauthorblockA{\textit{Department of Computer Science} \\
\textit{Kennesaw State University}\\
Marietta, USA \\
mkhan74@kennesaw.edu}
\and
\IEEEauthorblockN{Kazi Aminul Islam}
\IEEEauthorblockA{\textit{Department of Computer Science} \\
\textit{Kennesaw State University}\\
Marietta, USA \\
kislam4@kennesaw.edu}
}

\maketitle

\begin{abstract}

Federated learning enables distributed training without requiring clients to share their raw data. However, its reliance on the integrity of the client-submitted updates exposes the global model to stealthy backdoor poisoning. Existing defenses often rely on individual evidence sources, but stealth-constrained attacks can adapt to these signals. Such attacks can suppress anomaly signals they are optimized to evade, yet their poisoned updates still leave residual structural traces. We propose FedMAST, a Federated Multi-Axis Structural Tracing defense for backdoor detection in federated learning. FedMAST scores client updates using complementary structural, spectral, and historical evidence and then applies tiered filtering and round-level containment to limit adversarial influence. To capture traces that isolated signals may miss, FedMAST uses squeeze-pair coherence scoring to expose coupled feature distortions and signed spectral-drift tracking to reveal persistent directional changes over time. Across six backdoor attacks, FedMAST achieves lower attack success rate (ASR) than baseline defenses in all nine evaluated comparisons, averaging 1.51\% ASR and 94.84\% main-task accuracy (MTA) across the complete 200-round runs. Over the full 200-round method-aware \textsc{CovertLayers} run, FedMAST achieves 1.53\% ASR and 92.26\% MTA, compared with ASRs of 100.00\%, 99.67\%, 99.53\%, and 32.84\% for FedAvg, MultiKrum, AlignIns, and FLAME, respectively.

\end{abstract}


\section{Introduction}

Federated learning (FL) facilitates collaborative training for machine learning models using decentralized data sources~\cite{mcmahan2017communication}. However, malicious clients can embed a hidden behavior in the global model while preserving normal performance, making backdoor poisoning difficult to detect under regular evaluation settings~\cite{bhagoji2019analyzing,bagdasaryan2020backdoor}. Robust aggregation methods such as Krum, coordinate-wise median, and trimmed mean reduce the influence of updates that appear anomalous under distance or coordinate statistics~\cite{blanchard2017machine,yin2018byzantine}. However, their selectivity is limited in non-independent and identically distributed (non-IID) settings~\cite{hsu2019measuring}. Moreover, these methods can be bypassed when malicious updates are constrained to remain close to benign updates~\cite{guerraoui2018hidden,baruch2019little}.

Recent attacks improve stealth by distributing trigger components across
clients, restricting poisoning to selected parameters or layers, coordinating
client roles, or aligning malicious updates with benign
directions~\cite{xie2019dba,zhang2022neurotoxin,li20233dfed,zhuang2024backdoor,yang2025stealthy}. Existing defenses use strategies such as clustering, clipping, historical consistency, model inspection, direction alignment, and layer-aware filtering~\cite{nguyen2022flame,rieger2022deepsight,zhang2022fldetector,xu2025detecting,abacha2026fedsurrogate}. These signals are useful, but a defense centered on one evidence view can fail when the attacker explicitly optimizes against that view~\cite{bagdasaryan2020backdoor,xie2019dba,zhang2022neurotoxin,li20233dfed,zhuang2024backdoor,yang2025stealthy}. We observe that suppressing one anomaly signal while preserving an effective backdoor can shift the malicious footprint into another structural, historical, temporal, or spectral channel. We propose FedMAST, a Federated Multi-Axis Structural Tracing defense for stealth-constrained backdoor attacks. FedMAST evaluates the coherence across complementary evidence sources and limits the influence of suspicious updates.

The contributions of this work are as follows.

\begin{itemize}

\item \textbf{FedMAST: Federated Multi-Axis Structural Tracing.}
We propose a server-side defense that scores client updates across complementary structural, historical, and spectral evidence and enforces round-level containment and defense-state quarantine to limit adversarial influence.

\item \textbf{Signed temporal spectral-drift features.}
We introduce stage-resolved spectral features that accumulate signed changes in singular-value concentration and spectral entropy across repeated appearances of each client partition. Their separation from the unsigned structural bank preserves attack-direction information, exposing weak but persistent gradient-aligned drift.

\item \textbf{Squeeze-pair coherence scoring.}
We introduce a paired-feature anomaly axis that evaluates joint coherence across structurally related properties, exposing deviations that remain obscured when metrics are evaluated independently.

\item \textbf{Method-aware adaptive evaluation.} We construct \textsc{CovertLayers}, which extends BC-Layers by preserving its layer-critical backdoor objective while constraining the magnitude, directional, and layer-concentration evidence observable to FedMAST.

\end{itemize}

\section{Related Work}

\noindent\textbf{Federated backdoor attacks.}
Model-replacement attacks amplify malicious updates to steer the aggregated model toward an attacker-chosen adversarial behavior~\cite{bagdasaryan2020backdoor}. Attack of the Tails and Neurotoxin improve backdoor persistence against benign training dynamics~\cite{wang2020attack,zhang2022neurotoxin}. Distributed Backdoor Attack (DBA) distributes partial trigger patterns across coordinated clients to create an effective backdoor after aggregation~\cite{xie2019dba}. Recent attacks make the poisoning footprint more selective and defense-aware. 3DFed combines coordinated roles, noise masks, decoy models, and adaptive tuning to evade multiple defense classes~\cite{li20233dfed}. BC-Layers concentrates poisoning in backdoor-critical layers~\cite{zhuang2024backdoor}. Layer-wise gradient alignment (LGA) aligns malicious updates with benign layer-wise directions~\cite{yang2025stealthy}. Together, these attacks show that distance, direction, layer concentration, and round-local clustering can become explicit evasion targets.

\noindent\textbf{Defenses against model poisoning and backdoors.}
Krum and MultiKrum select updates using neighbor-distance criteria, while the coordinate-wise median and trimmed mean reduce the effect of extreme coordinate values~\cite{blanchard2017machine,yin2018byzantine}. FLTrust uses trusted server data, FLDetector uses historical prediction, and CRFL constrains global-model movement~\cite{cao2020fltrust,zhang2022fldetector,xie2021crfl}. These approaches strengthen aggregation, but each still emphasizes a specific source of evidence or auxiliary assumption. Complementing these approaches, backdoor-specific defenses introduce evidence sources tailored to poisoned behavior. FLAME combines clustering, clipping, and noise injection, while DeepSight uses model-behavior inspection and multi-metric defenses combine several update-level indicators~\cite{nguyen2022flame,rieger2022deepsight,huang2023multi}. Other methods use frequency information, direction alignment, or layer criticality with surrogate replacement~\cite{fereidooni2023freqfed, xu2025detecting,abacha2026fedsurrogate}. Stateful defenses incorporate cross-round information in backdoor detection and mitigation~\cite{ali2024adversarially,wan2026mars}. These defenses provide useful evidence sources, but each source can also become an evasion target for a defense-aware adaptive attacker. 

FedMAST differs from prior defenses by evaluating coherence among anomaly signals and how they evolve across rounds, rather than relying on a single aggregate score. Its signed temporal spectral-drift trace preserves the direction of stage-wise spectral changes across repeated client appearances, while squeeze-pair coherence explicitly models relationships between structurally related features~\cite{huang2023multi,zhang2026heterogeneity}. This design targets residual inconsistencies that may persist when an adaptive attacker suppresses magnitude, directional, spectral, or layer-wise anomaly signals.

\section{Proposed Methodology}

\subsection{Problem Setup and Threat Model}
\label{sec:threat-model}

\begin{figure}[!htbp]
    \centering
    \includegraphics[width=\linewidth]{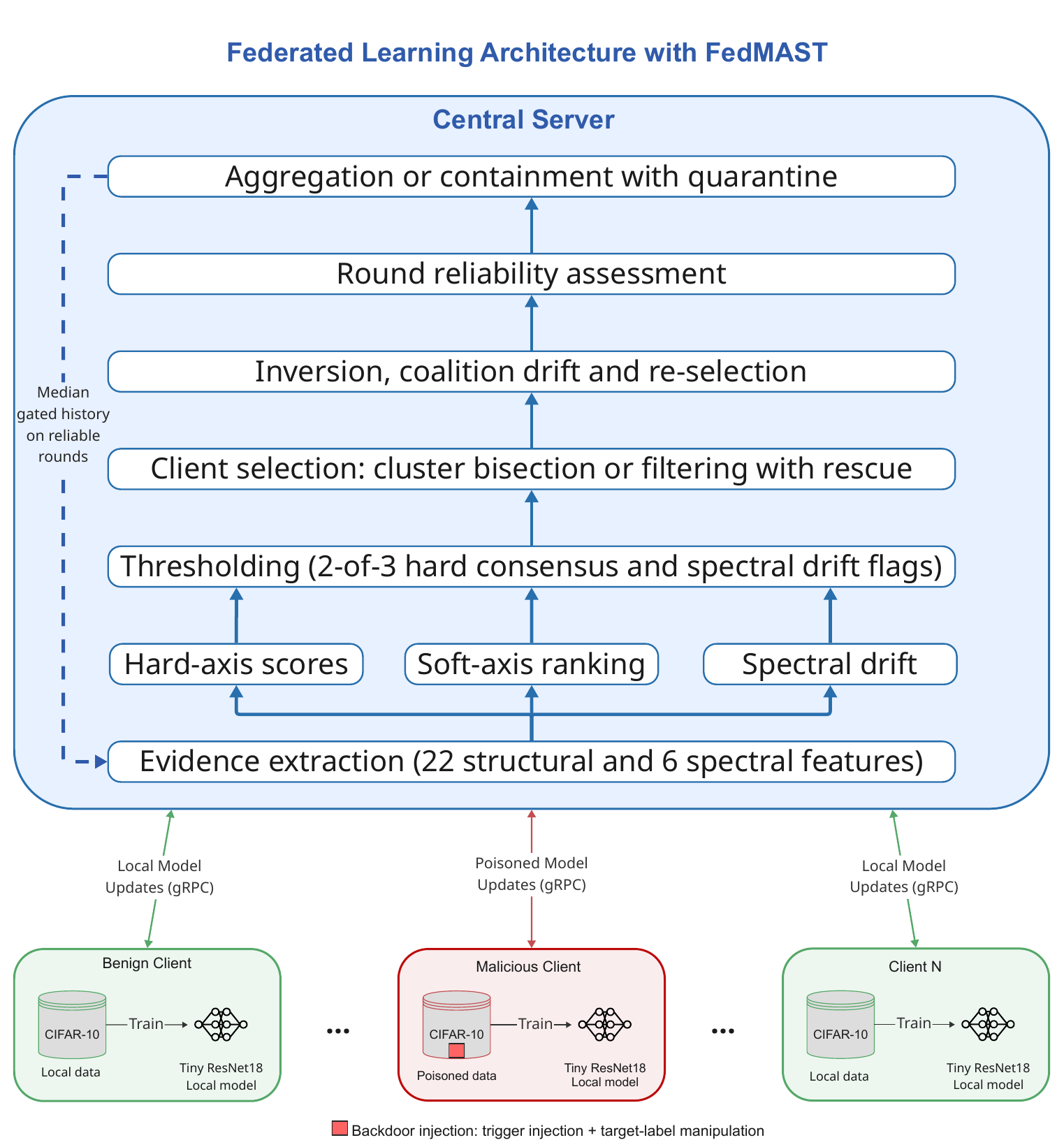}
    \caption{FedMAST decision pipeline. Client updates are mapped to the 22-feature structural bank and six segregated spectral-drift features (Table~\ref{tab:feature-bank}), scored by Algorithm~\ref{alg:scoring}, and classified by the hard-consensus and spectral rules in Algorithm~\ref{alg:flagging}. Algorithm~\ref{alg:containment} constructs the accepted set and applies the anchor-relative cluster bisection, the accepted set inversion, and the coalition-drift validation. Reliable rounds use FedAvg and suspicious rounds use robust containment and quarantine the history update.}
    \label{fig:fl_architecture}
\end{figure}

We consider a federated learning setup that maintains a global model $w_t$ at communication round $t$. In the communication round $t$, the server samples a set of clients $S_t$, distributes the current global model $w_t$, and receives locally trained models $\{\widetilde{w}_{i,t}\}_{i\in S_t}$. The update submitted by the client $i$ is \begin{equation} \Delta_{i,t}=\widetilde{w}_{i,t}-w_t. \end{equation} The server applies an aggregation rule $\mathcal{A}_t$ to produce the next global model. It observes the local model updates and protocol-level metadata, but does not observe raw client data and does not require a trusted reference dataset.

A subset $M_t \subset S_t$ may be controlled by an attacker. Malicious
clients may poison their local data, modify local training, coordinate their behavior, or post-process their local model updates. Their objective is to induce a trigger-conditioned target behavior in the global model while preserving the model utility and suppressing observable evidence sufficiently to remain admissible under the server's aggregation and defense rules. A method-aware attacker may adapt against the observable structure of FedMAST, but does not observe the current benign updates or the server's runtime historical state, thresholds, buffers, and spectral accumulators. FedMAST requires stable pseudonymous client identifiers but does not require clean server-side data. It can initialize from a frozen history or build rolling history online from low-risk accepted updates. Let $p_i$ denote client $i$'s stable pseudonymous partition identifier, and let $\mathcal D=\{\mathrm{round},\mathrm{squeeze},\mathrm{hist},
\mathrm{anchor},\mathrm{traj},\mathrm{mom},\mathrm{spec}\}$ denote the seven evidence axes.
 
\begin{algorithm}[!htbp]
\DontPrintSemicolon
\SetKwInOut{Input}{Input}
\SetKwInOut{Output}{Output}
\SetKwInOut{Param}{Param}
\Input{Global model $w_t$; submitted models
       $\{\tilde{w}_{i,t}\}_{i \in S_t}$; historical state
       $\mathcal{H}_{t-1}$; $\{p_i\}_{i\in S_t}$}
\Param{Spectral Exponential Moving Average (EMA) decay $\alpha$, catastrophe threshold $\theta_{\mathrm{cat}}{=}100$,
       squeeze pairs $\mathcal{Q}$}
\Output{Scorable set $S_t'$;
        scores $\{D_d(i)\}_{d\in\mathcal D,\,i\in S_t'}$;
        anchor-validity indicator $\mathit{anc\_ok}$;
        $\{\mathrm{RankScore}(i)\}_{i\in S_t'}$}
Extract structural feature vector $\mathbf{x}_i=(x_{i,k})_k$ and spectral features $\forall\, i \in S_t$\;
$S_t' \leftarrow$ clients with valid feature extraction\;
Retrieve historical baselines and temporal/spectral states from
$\mathcal H_{t-1}$\;
Compute structural $z_{i,k}$ using Eq.~\eqref{eq:robust_z}\;
Compute $D_{\mathrm{round}},D_{\mathrm{squeeze}},
D_{\mathrm{hist}},D_{\mathrm{anchor}}$ using
Eqs.~\eqref{eq:dround}, \eqref{eq:dsq},
\eqref{eq:dhist}, and \eqref{eq:danchor}\;
Normalize trajectory statistics and momentum to
$z^{\mathrm{traj}}_{i,h}$ and $z^{\mathrm{mom}}_i$;
compute $D_{\mathrm{traj}},D_{\mathrm{mom}}$ using
Eqs.~\eqref{eq:dtraj} and \eqref{eq:dmom}\;
Normalize spectral features to $\hat z_{i,k}$ using Eq.~\eqref{eq:robust_z}\;
\ForEach{$i\in S_t'$ with valid spectral evidence}{
  Compute $s_{i,t}$ and $D_{\mathrm{spec}}(p_i)$ using Eqs \eqref{eq:signed_spectral} and \eqref{eq:spectral_ema}\;
}
\tcp{Anchor validation and ranking}
$\mathit{anc\_ok}\leftarrow
\mathrm{med}_{i\in S_t'}D_{\mathrm{anchor}}(i)
\leq\theta_{\mathrm{cat}}$\;
\lIf{$\neg\mathit{anc\_ok}$}{
  set $D_{\mathrm{anchor}}(\cdot)\leftarrow0$}
$\mathrm{RankScore}(i)\leftarrow
\max_{d\in\mathcal D}D_d(i)$\;
\Return $S_t',\,\{D_d(i)\},\,\mathit{anc\_ok},\,
\{\mathrm{RankScore}(i)\}_{i\in S_t'}$\;
\caption{FedMAST Evidence Scoring}
\label{alg:scoring}
\end{algorithm}

\subsection{Multi-Axis Evidence Extraction}

\begin{table}[!t]
\centering
\caption{FedMAST structural feature bank.}
\label{tab:feature-bank}
\scriptsize
\setlength{\tabcolsep}{2.0pt}
\renewcommand{\arraystretch}{0.98}
\begin{tabular}{@{}
>{\raggedright\arraybackslash}p{0.20\linewidth}
>{\raggedright\arraybackslash}p{0.72\linewidth}
@{}}
\toprule
\textbf{Family} & \textbf{Features and related evidence} \\
\midrule

Global update geometry &
Update $\ell_2$ norm, $\ell_2$ distance to round mean,
$\ell_2$ distance to historical baseline, cosine to round mean
~\cite{bagdasaryan2020backdoor,nguyen2022flame}. \\

Directional
alignment &
Cosine to leave-one-out mean, cosine to historical baseline,
stage-norm-signature cosine to round mean, head-sign agreement with
historical baseline
~\cite{cao2020fltrust,xu2025detecting}. \\

Layer-energy
conservation &
Head $\ell_2$ norm, layer-4 $\ell_2$ norm, head/total norm ratio,
classifier/backbone norm ratio, head/total norm ratio $\times$ maximum backbone
kurtosis
~\cite{zhuang2024backdoor,yang2025stealthy,abacha2026fedsurrogate}. \\

Spectral /
shape &
Head spectral entropy, head top-SV ratio, layer-4 skewness,
layer-3 kurtosis, maximum backbone kurtosis
~\cite{fereidooni2023freqfed,rieger2022deepsight}. \\

Cross-layer
consistency &
Layer-4 cosine to round mean, head cosine to leave-one-out mean,
classifier-row update-magnitude entropy, layer-4 $\ell_\infty/\ell_2$ ratio
~\cite{huang2023multi,abacha2026fedsurrogate}. \\

\bottomrule
\end{tabular}
\end{table}

Figure~\ref{fig:fl_architecture} summarizes the FedMAST pipeline. Each update is represented by the 22-feature structural bank in Table~\ref{tab:feature-bank}. Six additional layer-2--4 top-SV-ratio and spectral-entropy features are reserved for signed temporal $D_{\mathrm{spec}}$ scoring, preserving their directional information. FedMAST normalizes each structural-bank feature $k$ within the current round using
\begin{equation}
z_{i,k} =
\frac{x_{i,k}-\operatorname{med}_{j\in S_t'}x_{j,k}}
{1.4826\,\operatorname{MAD}_{j\in S_t'}(x_{j,k})}.
\label{eq:robust_z}
\end{equation}

where $x_{i,k}$ denotes the value of structural feature $k$ for client
$i$, and $j$ indexes clients in the scorable set $S_t'$. Non-finite feature values are replaced by the median of the finite round values and a feature with no finite values or scaled MAD below $10^{-12}$ contributes zero. FedMAST then computes
\begin{equation}
D_{\mathrm{round}}(i)
=\max_{f\in\mathcal F}
\operatorname{top2mean}_{k\in f}|z_{i,k}|,
\label{eq:dround}
\end{equation}
where $\mathcal F$ contains the five families in Table~\ref{tab:feature-bank} and $D_d(i)$ denotes client $i$'s score on axis $d\in\mathcal D$.

\subsection{Squeeze-Pair Tracing}

FedMAST uses squeeze-pair tracing to capture the evasion tension between coupled evidence signals. For a fixed set of squeeze pairs $\mathcal{Q}$, FedMAST computes
\begin{equation}
D_{\mathrm{squeeze}}(i)
=
\max_{(a,b)\in \mathcal{Q}}
\sqrt{z_{i,a}^{2} + z_{i,b}^{2}}.
\label{eq:dsq}
\end{equation}

Here, $a$ and $b$ index the two features in each squeeze pair $(a,b)\in\mathcal{Q}$ whose relationship is expected to remain coherent for benign updates. The squeeze pairs are derived from structural constraints. The distance to the historical baseline is paired with the layer-4 $\ell_2$ norm to expose updates that suppress global displacement while concentrating change in late layers. The cosine-to-round mean is paired with the head spectral entropy to test whether benign-like directional alignment also preserves classifier-head structure. The head/total norm ratio is paired with the head top-SV ratio to couple energy allocation with spectral concentration. Finally, the classifier/backbone norm ratio is paired with maximum backbone kurtosis to expose localized backbone distortion despite a benign-like layer-energy balance. A stealth-constrained attacker may suppress one feature, but doing so can shift residual evidence to its paired counterpart.

\subsection{Historical and Temporal Context}

Single-round statistics can be distorted by non-IID heterogeneity or coordinated malicious clients. To mitigate this, FedMAST computes $D_{\mathrm{hist}}$ by re-standardizing structural evidence against a frozen or rolling baseline $\mathcal{B}$ of previously accepted updates. Its threshold uses a second MAD pass after removing first-pass outliers and is bounded below by half the first-pass threshold; we denote this
rule by $\mathrm{IterMAD}(\cdot,k)$. The historical score is

\begin{equation}
D_{\mathrm{hist}}(i)=
\max_f \operatorname{top2mean}_{k\in f}
\left|
\frac{z_{i,k}-m^{z}_{\mathcal{B},k}}
{s^{z}_{\mathcal{B},k}}
\right|.
\label{eq:dhist}
\end{equation}

While $D_{\mathrm{hist}}$ measures the historical inconsistency per-client, $D_{\mathrm{anchor}}$ measures raw-feature distance from the accepted historical center and is used for cluster bisection, inversion checks, and anchor-based re-selection. FedMAST disables anchor-dependent decisions when the sampled population is collectively far from this reference. Together, they distinguish relative disagreement within the current round from absolute deviation from the trusted historical structure.

The raw historical anchor is
\begin{equation}
D_{\mathrm{anchor}}(i)=
\max_f \operatorname{top2mean}_{k\in f}
\left|
\frac{x_{i,k}-m^{x}_{\mathcal{B},k}}
{s^{x}_{\mathcal{B},k}}
\right|.
\label{eq:danchor}
\end{equation}

Here, $m_{\mathcal{B},k}$ and $s_{\mathcal{B},k}$ are historical medians
and scaled MADs for normalized ($z$) and raw ($x$) features. Features without a valid historical MAD do not contribute to the corresponding score. Client trajectories track the classifier/backbone ratio, head/total ratio, head top-SV ratio, layer-4 $\ell_\infty/\ell_2$ ratio, and layer-4 skewness using EMA deviation (weight $0.3$), volatility, absolute linear trend, and mean consistency of successive changes. Ranking momentum is an EMA (decay $0.7$) of prior RankScore percentiles and activates after five appearances. Both scores require sufficient history and affect ranking without independently triggering rejection.

\begin{align}
D_{\mathrm{traj}}(i)
&=\operatorname{top2mean}_{h}
  |z^{\mathrm{traj}}_{i,h}|,
\label{eq:dtraj}\\
D_{\mathrm{mom}}(i)
&=\max\{0,z^{\mathrm{mom}}_i\}.
\label{eq:dmom}
\end{align}

\begin{algorithm}[t]
\DontPrintSemicolon
\SetKwInOut{Input}{Input}
\SetKwInOut{Output}{Output}
\SetKwInOut{Param}{Param}
\Input{Scorable set $S_t'$; axis scores from Alg.~\ref{alg:scoring};
       spectral states $D_{\mathrm{spec}}(p)$ and appearance counts
       $\mathrm{app}(p)$}
\Param{MAD multiplier $k$, consensus $c{=}2$,
       spectral percentile $q$ with operator $P_q$, min appearances $n_a$,
       spectral floor $\theta_{\min}$}
\Output{Hard-consensus set $H$; extreme set $H^{*}$;
        spectral set $\Sigma$; anchor threshold $\theta_{\mathrm{anchor}}$}
\tcp{Per-axis thresholding}
$\mathcal{D}_{\mathrm{hard}} \leftarrow
  \{\mathrm{round},\mathrm{squeeze},\mathrm{hist}\}$\;
\ForEach{$d \in \{\mathrm{round},\mathrm{squeeze},\mathrm{anchor}\}$}{
  $\theta_d \leftarrow  
    \mathrm{med}_{i \in S_t'}D_d(i)
    + k \cdot 1.4826 \cdot
      \mathrm{MAD}_{i \in S_t'}D_d(i)$\;
}
$\theta_{\mathrm{hist}} \leftarrow
  \mathrm{IterMAD}\!\left(
    \{D_{\mathrm{hist}}(i)\}_{i \in S_t'},\, k
  \right)$\;
$\theta_{\mathrm{spec}} \leftarrow
  \max\!\bigl(
    P_q\!\bigl(\{D_{\mathrm{spec}}(p) :
      \mathrm{app}(p) \geq n_a\}\bigr),\;
    \theta_{\min}
  \bigr)$\;
\lIf{fewer than five partitions have $\mathrm{app}(p)\geq n_a$}{
  $\theta_{\mathrm{spec}}\leftarrow\infty$}  
$F_d \leftarrow \{i \in S_t' : D_d(i) > \theta_d\}
  \quad \forall\, d \in \mathcal{D}_{\mathrm{hard}}$\;
\BlankLine

\tcp{Client classification}
$v_i\leftarrow\sum_{d\in\mathcal D_{\mathrm{hard}}}
  \mathbf{1}[i\in F_d]\quad\forall i\in S_t'$\;
$H\leftarrow\{i\in S_t':v_i\geq c\}$\;
$H^*\leftarrow\{i\in S_t':v_i=3,\;
  \exists d\in\mathcal D_{\mathrm{hard}}:
  D_d(i)>2\theta_d\}$\;
$\Sigma\leftarrow\{i\in S_t':
  \mathrm{app}(p_i)\geq n_a,\;
  D_{\mathrm{spec}}(p_i)>\theta_{\mathrm{spec}}\}$\;
\Return $H,\,H^*,\,\Sigma,\,\theta_{\mathrm{anchor}}$\;
\caption{FedMAST Thresholding and Flagging}
\label{alg:flagging}
\end{algorithm}

\subsection{Signed Temporal Spectral Drift}

Gradient-aligned attacks can remain close to benign update directions while gradually altering the spectral structure of the backbone stages~\cite{fereidooni2023freqfed,yang2025stealthy}. FedMAST computes stage-resolved singular-value concentration and spectral entropy features, then accumulates their signed drift across repeated appearances of each client partition. Spectral evidence is valid when at least two top-SV and two entropy features are available. After normalizing the spectral features, the signed appearance score is

\begin{equation}
s_{i,t}
=
-\overline{\hat{z}}_{\mathrm{topSV}}(i,t)
+
\overline{\hat{z}}_{\mathrm{entropy}}(i,t).
\label{eq:signed_spectral}
\end{equation}

For each appearance $a$ of partition $p_i$, FedMAST accumulates its signed spectral score as
\begin{equation}
D_{\mathrm{spec}}^{(a)}(p_i)=
\begin{cases}
s_{i,t}, & a=1,\\
\alpha D_{\mathrm{spec}}^{(a-1)}(p_i)
 +(1-\alpha)s_{i,t}, & a>1.
\end{cases}
\label{eq:spectral_ema}
\end{equation}

Here $\alpha$ is the spectral moving-average decay. Spectral decisions are enabled only after sufficient appearances.

\subsection{Client Scoring and Round-Level Containment}

\begin{figure}[t]
    \centering
    \includegraphics[width=\columnwidth,trim=6 6 6 6,clip]{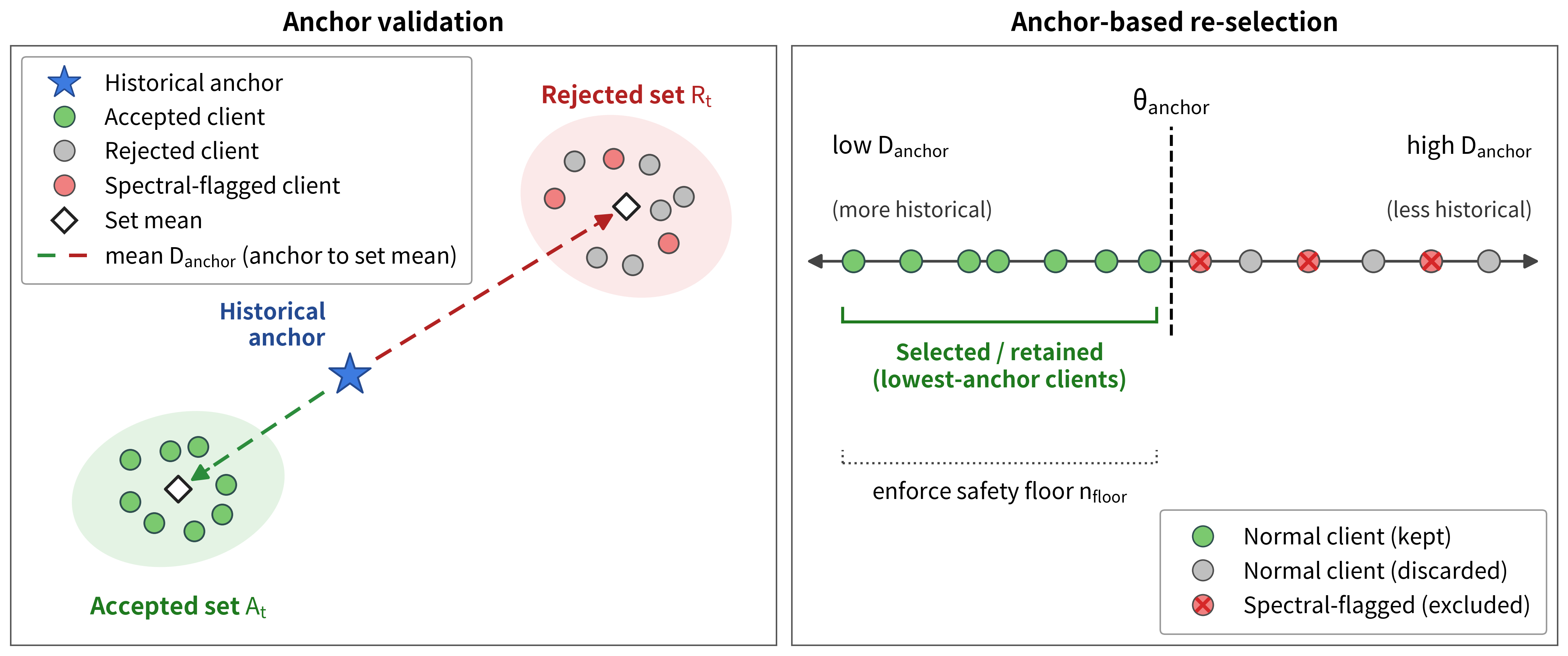}
    \caption{(a) In a normal round the accepted set lies nearer the historical anchor than the rejected set and the round is flagged as inverted if the accepted set's mean anchor distance exceeds the rejected set's. (b) Re-selection ranks clients by anchor distance, excludes spectral-flagged clients, keeps at least $n_{\mathrm{floor}}$, and admits further clients up to $\theta_{\mathrm{anchor}}$.}
    \label{fig:anchor_validation}
\end{figure}

The hard axes use 2-of-3 consensus, while $\Sigma$ is an independent spectral flag.  Cluster bisection standardizes raw feature vectors against the historical raw median and MAD, initializes two clusters from the farthest pair, and accepts a split only when both clusters contain at least two clients and the farther centroid is at least $1.5\times$ as distant from the historical origin as the nearer one~\cite{nguyen2022flame}. FedMAST then validates the accepted coalition using accepted-set inversion and aggregate drift. Aggregate drift $D_{\mathrm{agg}}(A_t)$ is the cosine distance between the accepted coalition's mean six-stage $\ell_2$-norm signature and its historical EMA with decay $0.8$, activated after five reliable rounds and flagged above $0.3$. A failed validation marks the round as suspicious and triggers anchor-based re-selection or containment. Reliable rounds use FedAvg on the accepted set. Anchor re-selection (Fig.~\ref{fig:anchor_validation}) ranks clients by $D_{\mathrm{anchor}}$, excludes $\Sigma$, admits enough clients to satisfy $n_{\mathrm{floor}}$, and admits additional clients only when $D_{\mathrm{anchor}}\leq\theta_{\mathrm{anchor}}$. Suspicious rounds invoke the configured containment policy $\mathrm{Contain}_{\pi}$ (no-op, filtered FedAvg, coordinate median, or trimmed mean), which retains $w_t$ when $|A_t|<n_{\mathrm{floor}}$.

\begin{algorithm}[!htbp]
\DontPrintSemicolon
\SetKwInOut{Input}{Input}
\SetKwInOut{Output}{Output}
\SetKwInOut{Param}{Param}

\Input{$S_t'$, $D_{\mathrm{anchor}}$, $\mathrm{RankScore}$,
       $\mathit{anc\_ok}$ from Alg.~\ref{alg:scoring};
       $H,H^*,\Sigma,\theta_{\mathrm{anchor}}$ from
       Alg.~\ref{alg:flagging};
       $w_t$, $\{\tilde{w}_{i,t}\}_{i\in S_t'}$, $\mathcal H_{t-1}$}
\Param{$n_{\min}{=}5$, $n_{\mathrm{floor}}{=}3$,
       containment policy $\pi$}
\Output{$w_{t+1}$, $\mathcal H_t$}

\tcp{Client decision}
$\mathit{susp}\leftarrow\mathtt{false}$\;

\eIf{$\mathit{anc\_ok}$ \textup{ and bisection selects }
     $\mathcal C_{\mathrm{near}}$}{
  $A_t\leftarrow
    \mathcal C_{\mathrm{near}}\setminus(\Sigma\cup H^*)$\;
  eject highest-$\mathrm{RankScore}$ clients in $H\cap A_t$
    while $|A_t|>n_{\mathrm{floor}}$\;
  $\mathit{susp}\leftarrow\mathtt{true}$\;
}{
  $A_t\leftarrow S_t'\setminus(H\cup\Sigma)$\;
  \If{$|A_t|<n_{\min}$}{
    rescue lowest-$\mathrm{RankScore}$ clients from
    $H\setminus(\Sigma\cup H^*)$
    until $|A_t|=n_{\min}$ or none remain\;
    $\mathit{susp}\leftarrow\mathtt{true}$\;
  }
}

\tcp{Historical validation}
$\begin{aligned}
\mathrm{Inv}\equiv{}&
|A_t|\ge3,\quad |S_t'\setminus A_t|\ge2,\\[-1pt]
&\overline D_{\mathrm{anc}}(A_t)>
\max\!\left\{10,\,
1.5\,\overline D_{\mathrm{anc}}(S_t'\setminus A_t)\right\}
\end{aligned}$\;

$\begin{aligned}
\mathrm{Drift}\equiv{}&
\textup{no split}\land(\textup{number of reliable rounds}\ge5)\\[-1pt]
&\land |A_t|\ge3\land D_{\mathrm{agg}}(A_t)>0.3
\end{aligned}$\;

\If{$\mathit{anc\_ok}\land(\mathrm{Inv}\lor\mathrm{Drift})$}{
  $A_t\leftarrow$ anchor re-selection using
    $\theta_{\mathrm{anchor}}$\;
  $\mathit{susp}\leftarrow\mathtt{true}$\;
}

$\mathit{susp}\leftarrow
  \mathit{susp}\lor|\Sigma|>0$\;

\tcp{Aggregation}
\eIf{$\neg\mathit{susp}$}{
  $w_{t+1}\leftarrow
    \mathrm{FedAvg}
    (\{\tilde w_{i,t}:i\in A_t\})$\;
  $\mathcal H_t\leftarrow$ admit
    $\{i\in A_t:
      \mathrm{RankScore}(i)\le
      \operatorname{med}_{j\in S_t'}
      \mathrm{RankScore}(j)\}$\;
}{
  $w_{t+1}\leftarrow
    \mathrm{Contain}_{\pi}
    (\{\tilde w_{i,t}:i\in A_t\},w_t)$
    \tcp*{returns $w_t$ if $|A_t|<n_{\mathrm{floor}}$}
  $\mathcal H_t\leftarrow
    \mathcal H_{t-1}$ \tcp*{quarantine}
}

\Return $w_{t+1},\mathcal H_t$\;
\caption{FedMAST Aggregation}
\label{alg:containment}
\end{algorithm}

\subsection{FedMAST-Aware Adaptive Attack}

We construct \textsc{CovertLayers}, a method-aware extension of
BC-Layers~\cite{zhuang2024backdoor}. Its objective is to preserve the
trigger-critical layer modifications responsible for the backdoor while reducing the magnitude, direction, and layer-concentration evidence visible to FedMAST. Following the method-aware setting in Sec.~\ref{sec:threat-model}, the attacker estimates the benign update structure using local proxy models and derives the critical-layer set $\mathcal{L}^{*}$ through layer-substitution analysis. For each critical layer $\ell\in\mathcal{L}^{*}$, the attacker softens the malicious replacement through

\begin{equation}
\tilde{w}_{\ell}
=
\lambda w_{\ell}^{\mathrm{mal}}
+
(1-\lambda)\bar{w}_{\ell}^{\mathrm{ben}},
\qquad 0 < \lambda \leq 1,
\end{equation}
where $\bar{w}^{\mathrm{ben}}$ is the benign proxy average and $w^{\mathrm{mal}}$ is the malicious model. Non-critical layers are initialized from the benign reference. The attacker then applies three structural evasion steps in the trainable-parameter space. First, it injects calibrated noise into non-critical trainable layers to smooth the layer-energy profile and avoid an update that appears surgically concentrated only in $\mathcal{L}^{*}$. Second, it blends the crafted update direction toward the mean benign proxy update,

\begin{equation}
\Delta' =
(1-\beta)\Delta
+
\beta \bar{\Delta}^{\mathrm{ben}},
\end{equation}

where $\Delta$ is the malicious update crafted and $\bar{\Delta}^{\mathrm{ben}}$ is the mean proxy update. The attacker also projects its norm toward the median benign proxy norm. If the post-evasion backdoor success falls below a minimum fraction of the full malicious model's success, the attacker relaxes the evasion constraints in stages. This reduces directional blending, reduces non-critical-layer smoothing, weakens norm matching, increases $\lambda$ toward a stronger BC-layer injection, and, if necessary, adds the next most critical layer to $\mathcal{L}^{*}$. 

\begin{figure*}[!t]
\centering
\includegraphics[width=\textwidth]{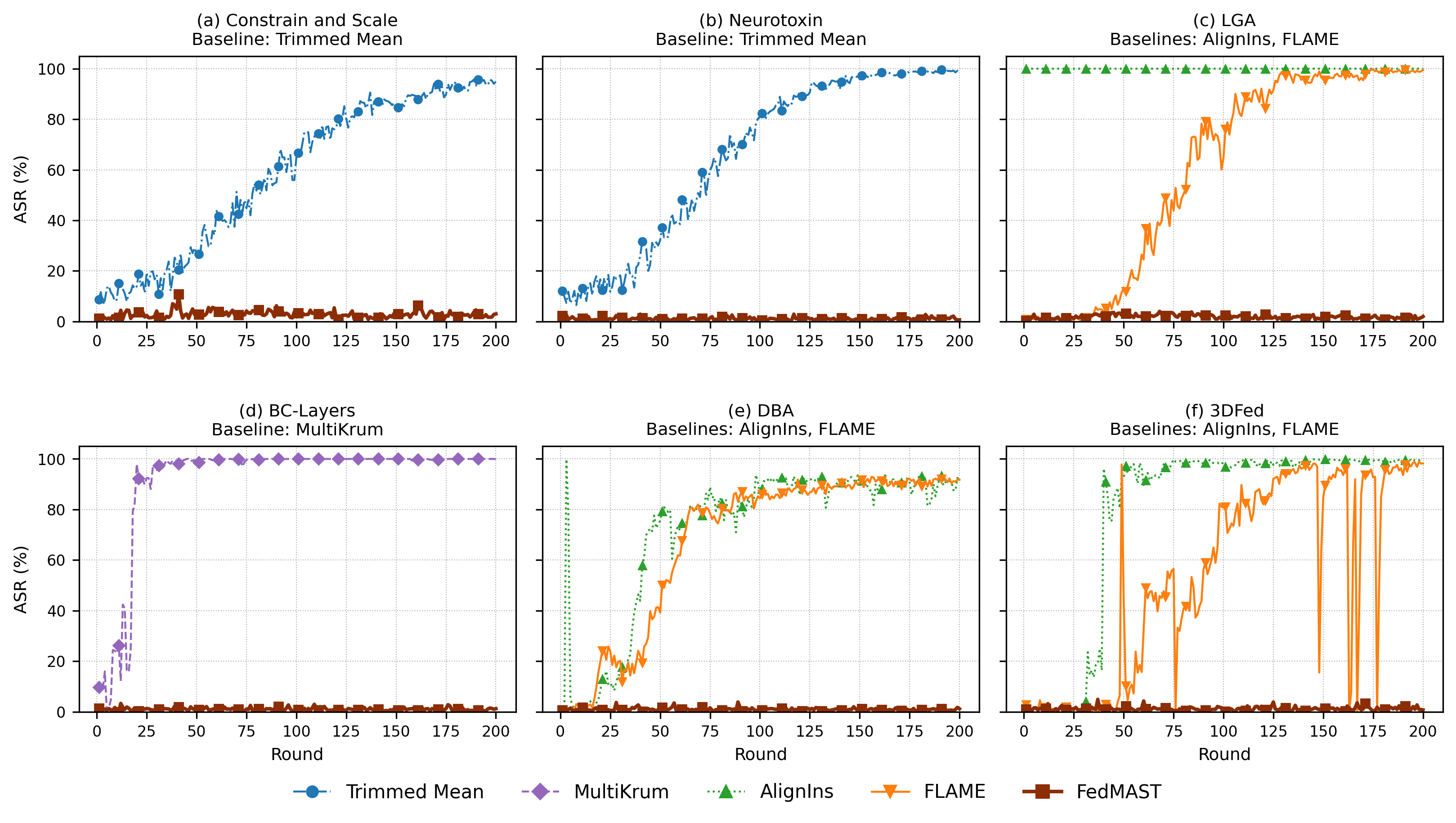}
\caption{ASR trajectories over 200 communication rounds for six stealth-constrained FL backdoor attacks. Each panel compares FedMAST with the baseline defenses for the corresponding attack.}
\label{fig:asr-containment}
\end{figure*}

\section{Experimental Setup}

\subsection{Federated Setup and Training}

We evaluate on CIFAR-10 with $N=100$ clients and 10 sampled clients per round for 200 communication rounds. Client data are partitioned using Dirichlet non-IID sampling with $\alpha=0.9$ and each partition uses an 80/20 local train-test split with seed 42. All experiments use a lightweight CIFAR-style Tiny ResNet-18 with approximately $0.27$M parameters. Benign clients train with stochastic gradient descent using momentum $0.9$, weight decay $5 \times 10^{-4}$, label smoothing $0.05$, batch size 64, and cosine learning-rate decay. To model natural client variability, benign clients independently sample the local learning rate from $\{0.003,0.004,0.005\}$ and the local epochs from $\{1,2,3\}$. Malicious clients start from the same global model as benign clients, but replace the local procedure with the attack-specific objective. Unless an attack requires a distinct trigger construction, attacks use target label 2 and fixed-trigger attacks use 20 triggered samples per poisoned minibatch. All defenses use identical sampling, partitions, and attack budgets. The number of malicious clients remains within the adversarial budget assumed by each defense. All primary CIFAR-10 evaluations initialize FedMAST from a frozen 50-round history checkpoint. The raw and normalized historical baselines are loaded read-only, so the two-round warmup is used only when no frozen history is available.  FedMAST uses the coordinate median as its containment policy for suspicious rounds in all experiments. FedMAST uses a MAD multiplier of $k=3$, a minimum accepted set of five clients, a two-round warmup, a 20-round history buffer, a 15-appearance trajectory window, spectral EMA decay $0.7$, a minimum of five appearances before spectral flagging, an 85th-percentile spectral threshold, and a spectral floor of $0.5$. Cluster bisection uses a centroid-distance ratio of $1.5$, and aggregate drift is flagged above $0.3$. These values were selected once as conservative operating defaults rather than optimized for any particular attack or dataset. We additionally evaluate the same FedMAST implementation on the 47-class EMNIST-Balanced split and FEMNIST to assess cross-dataset generalization. The implementation, attack and defense configurations, and scripts used to reproduce the CIFAR-10 experiments are archived in the accompanying artifact~\cite{fedmast_artifact}.

\subsection{Evaluation Metrics}

We centrally evaluate the global model after each communication round using main-task accuracy (MTA) on clean test samples and attack success rate (ASR) on trigger-poisoned test samples. The ASR is computed only on test samples whose true label is not the target label, preventing naturally target-class examples from inflating attack success. For DBA, ASR is measured using the assembled global trigger. The MTA and ASR in the results tables are reported as $\mu\pm\sigma$ over the last 100 rounds to characterize steady-state performance. 

\subsection{Baselines}

Table~\ref{tab:attack-selection} summarizes the attack suite and matched baseline defenses. Baselines are selected according to the attack mechanism they are designed to address, enabling focused comparisons across complementary threat settings.

\begin{table}[!htbp]
\centering
\caption{Attack and baseline-defense selection.}
\label{tab:attack-selection}
\scriptsize
\setlength{\tabcolsep}{3.2pt}
\renewcommand{\arraystretch}{1.08}
\begin{tabular}{@{}p{0.26\linewidth}p{0.42\linewidth}p{0.26\linewidth}@{}}
\toprule
Attack & Rationale & Baseline \\
\midrule
Constrain and Scale~\cite{bagdasaryan2020backdoor} & Scaling-based model replacement & Trimmed Mean~\cite{yin2018byzantine} \\
Neurotoxin~\cite{zhang2022neurotoxin} & Durable low-interference coordinates & Trimmed Mean \\
BC-Layers~\cite{zhuang2024backdoor} & Backdoor-critical layer concentration & MultiKrum~\cite{blanchard2017machine} \\
LGA~\cite{yang2025stealthy} & Layer-wise aligned stealth updates & \mbox{AlignIns~\cite{xu2025detecting}}, \mbox{FLAME~\cite{nguyen2022flame}} \\
DBA~\cite{xie2019dba} & Distributed trigger composition & AlignIns, FLAME \\
3DFed~\cite{li20233dfed} & Coordinated multi-role covert attack & AlignIns, FLAME \\
\bottomrule
\end{tabular}
\end{table}

\section{Results and Analysis}

\subsection{Defense Effectiveness}

\begin{table*}[!t]
\centering
\caption{Defense effectiveness against stealth-constrained FL backdoor attacks.}
\label{tab:main-defense-results}
\footnotesize
\setlength{\tabcolsep}{4.0pt}
\renewcommand{\arraystretch}{1.25}
\begin{tabular}{|l|l|c|c|c|c|c|c|c|c|c|}
\hline
\multirow{2}{*}{\textbf{Attack}} &
\multirow{2}{*}{\textbf{Baseline}} &
\multirow{2}{*}{\textbf{\# Mal.}} &
\multicolumn{4}{c|}{\textbf{Baseline defense}} &
\multicolumn{4}{c|}{\textbf{FedMAST}} \\
\cline{4-11}
 & & &
 MTA & ASR & Sel.\ (\%) & FPR (\%) &
 MTA & ASR & Sel.\ (\%) & FPR (\%) \\
\hline

C\&S & Trimmed Mean & 1 &
94.66$\pm$0.81 & 85.72$\pm$7.73 & -- & -- &
94.58$\pm$0.83 & 2.63$\pm$1.07 & 3.50 & 16.28 \\
\hline

Neurotoxin & Trimmed Mean & 1 &
94.47$\pm$0.95 & 94.37$\pm$5.22 & -- & -- &
94.52$\pm$0.97 & 1.10$\pm$0.51 & 0.00 & 10.44 \\
\hline

BC-Layers & MultiKrum & 1 &
82.89$\pm$3.26 & 99.97$\pm$0.10 & 57.00 & 50.78 &
94.46$\pm$0.84 & 1.01$\pm$0.52 & 0.00 & 15.44 \\
\hline

\multirow{2}{*}{LGA}
& AlignIns & 3 &
83.15$\pm$2.44 & 100.00$\pm$0.00 & 53.83 & 65.21 &
\multirow{2}{*}{95.76$\pm$0.76} &
\multirow{2}{*}{1.61$\pm$0.60} &
\multirow{2}{*}{7.73} &
\multirow{2}{*}{23.75} \\
\cline{2-7}
& FLAME & 3 &
95.08$\pm$0.84 & 94.69$\pm$5.77 & 50.29 & 26.16 &
& & & \\
\hline

\multirow{2}{*}{DBA}
& AlignIns & 4 &
75.88$\pm$4.20 & 90.17$\pm$2.68 & 18.88 & 54.00 &
\multirow{2}{*}{94.36$\pm$1.03} &
\multirow{2}{*}{0.93$\pm$0.41} &
\multirow{2}{*}{0.00} &
\multirow{2}{*}{4.23} \\
\cline{2-7}
& FLAME & 4 &
85.08$\pm$1.96 & 89.52$\pm$2.22 & 15.75 & 18.08 &
& & & \\
\hline

\multirow{2}{*}{3DFed}
& AlignIns & 3 &
92.63$\pm$2.61 & 99.01$\pm$0.88 & 28.07 & 51.78 &
\multirow{2}{*}{94.01$\pm$1.07} &
\multirow{2}{*}{1.17$\pm$0.74} &
\multirow{2}{*}{0.00} &
\multirow{2}{*}{9.82} \\
\cline{2-7}
& FLAME & 3 &
87.55$\pm$19.44 & 85.34$\pm$21.22 & 20.27 & 19.23 &
& & & \\
\hline

\multicolumn{11}{l}{\footnotesize
Sel.: percentage of malicious updates accepted; FPR: percentage of benign updates rejected; Mal.: Number of malicious clients per round.}\\
\multicolumn{11}{l}{\footnotesize
Sel. and FPR are reported over all applicable client appearances across the 200-round run. Not applicable for Trimmed Mean.}\\
\multicolumn{11}{l}{\footnotesize
MTA and ASR are reported as $\mu \pm \sigma$ over the final 100 communication rounds.}\\
\end{tabular}
\end{table*}

Figure~\ref{fig:asr-containment} shows round-wise ASR trajectories, while Table~\ref{tab:main-defense-results} reports final-100-round MTA and ASR. The baseline defenses leave sufficient malicious influence for the backdoor to persist. Across all six attacks, FedMAST suppresses persistent backdoor success while maintaining high MTA and low malicious selection.

\begin{table}[!htbp]
\centering
\caption{Effectiveness against an adaptive attacker.}
\label{tab:adaptive-attack-results}
\scriptsize
\setlength{\tabcolsep}{3.2pt}
\renewcommand{\arraystretch}{1.05}
\begin{tabular}{@{}lccc@{}}
\toprule
Method & MTA & ASR & Malicious Selection (\%) \\
\midrule
FedAvg & 85.72$\pm$1.63 & 99.76$\pm$0.20 & 100.00 \\
MultiKrum & 80.33$\pm$5.69 & 99.77$\pm$0.34 & 56.17 \\
AlignIns & 82.09$\pm$6.17 & 98.70$\pm$1.41 & 42.17 \\
FLAME & 90.43$\pm$6.45 & 26.23$\pm$16.38 & 40.62 \\
\textbf{FedMAST} & \textbf{94.77$\pm$0.93} & \textbf{0.95$\pm$0.49} & \textbf{0.00} \\
\bottomrule
\end{tabular}
\end{table}

\subsection{Effectiveness against adaptive attacker}

Table~\ref{tab:adaptive-attack-results} evaluates \textsc{CovertLayers}, which constrains structural traces utilized by FedMAST while preserving BC-Layers' layer-critical backdoor objective. The attack remains effective against FedAvg, MultiKrum, and AlignIns, while FLAME partially suppresses but does not eliminate it. FedMAST rejects all malicious updates in this setting and reduces the ASR below 1\% while preserving the highest MTA.

\subsection{False Positives and Detection Tradeoff}

Table~\ref{tab:main-defense-results} reports benign-rejection rates for FedMAST and client-selection baselines. Across the six experiments, five partitions accounted for 33.2\% of benign rejections despite only 6.5\% of benign appearances, with 69.2\% FPR versus 13.5\% overall. Under LGA ($\alpha=0.3$), five recurrent benign clients account for 28.26\% of false positives despite only 8.27\% of benign appearances, with a rejection rate of 82.98\% versus 19.00\% otherwise. Their normalized label entropy is 0.646 versus 0.620 overall and their dominant-class share is 43.3\% versus 45.9\%, indicating that their label distributions are broadly comparable to the client population. Despite this concentration, MTA changes only from 95.74\% before the attack window to 95.26\% in the final round.

\subsection{Sensitivity to Malicious Client Budget}

We stress FedMAST under increasing LGA attacker budgets, from 1 to 5 malicious clients among 10 sampled clients per round. The MTA remains stable across the sweep, ranging from 95.87\% to 96.19\%. ASR stays near the clean-trigger floor for 1--2 malicious clients at 1.27\% and 1.17\%, increases to 3.14\% with 3 malicious clients, and reaches 8.15\% when malicious clients constitute half of the sampled round. The final-round ASR follows the same trend, increasing from 0.71\% and 1.13\% at 1--2 attackers to 14.32\% at 5 attackers. 

\subsection{Runtime Overhead and Deployability}

FedMAST is server-side and does not change client training, communication, or model architecture. FedMAST adds 9.1\% end-to-end runtime overhead, increasing average execution from 98.5 to 107.4 minutes, or approximately 2.7 seconds per round.

\subsection{Ablation Study}
\label{ablationstudy} 
\noindent\textbf{Axis coverage.}
The axis coverage measures the fraction of malicious updates independently flagged by each FedMAST evidence axis. Neurotoxin, BC-Layers, and \textsc{CovertLayers} were detected almost uniformly by the structural and historical axes, with 99.5--100\% coverage across these three signals. LGA is detected primarily through the spectral axis (89.0\% versus \(\leq1.8\%\) on the others), while 3DFed shows the opposite pattern, with 100\% structural/historical coverage and 0\% spectral coverage. We pair each ablation with an attack that directly stresses the removed component.

\noindent\textbf{Component ablations.}
Table~\ref{tab:component-ablation} shows that LGA containment requires signed temporal spectral drift. Removing history infrastructure is less damaging than disabling spectral drift but still reduces the recall. For Constrain and Scale (C\&S) and DBA, removing an axis weakens consensus. This shows that the axes complement each other through both coverage and precision. We also evaluated a cold-start setting with (C\&S) active poisoning from round 1 and no frozen history. In this setting, malicious updates were initially accepted in the two warmup rounds, resulting in an ASR of 19.89\% and 20.50\% in rounds 1--2. From round 3 onward, FedMAST rejected all 98 subsequent malicious updates, yielding 100\% post-warmup recall (98\% overall).

\begin{table}[!htbp]
\centering
\caption{Component ablations over the attack window (\%).}
\label{tab:component-ablation}
\scriptsize
\setlength{\tabcolsep}{3.2pt}
\renewcommand{\arraystretch}{1.05}
\begin{tabular}{@{}llrrr@{}}
\toprule
Attack & Variant & ASR & Recall & FPR \\
\midrule
LGA & Full FedMAST                 & 1.18  & 99.67  & 23.00 \\
    & No spectral drift            & 24.14 & 29.63  & 24.97 \\
    & Spectral drift included in feature bank       & 25.91 & 23.87  & 23.65 \\
    & Unsigned per-round spectral  & 22.74 & 34.57  & 28.53 \\
    & No history infrastructure    & 5.70  & 67.49  & 18.17 \\
\midrule
C\&S & Full FedMAST               & 1.50  & 99.00  & 18.33 \\
     & No $D_{\mathrm{squeeze}}$   & 1.82  & 92.00  & 22.22 \\
\midrule
DBA & Full FedMAST                 & 1.09  & 100.00 & 0.00 \\
    & No $D_{\mathrm{hist}}$       & 1.07  & 100.00 & 9.10 \\
\bottomrule
\end{tabular}
\end{table}

\subsection{Heterogeneity and Cross-Dataset Generalization}

\noindent\textbf{Non-IID sensitivity.} 
We also evaluate FedMAST under LGA while varying the Dirichlet partition parameter $\alpha \in \{0.3,0.5,0.7\}$. FedMAST maintains stable MTA between 95.55\% and 95.93\% and limits ASR to 1.74--2.21\%. FPR remains the primary cost, ranging from 24.36\% to 29.37\%. False positives remain partition-concentrated across runs, with the top five false-positive partitions having an aggregate FPR of 81.8--92.6\%. 

\noindent\textbf{Cross-dataset generalization.}
We further evaluate the same FedMAST defense implementation on EMNIST-Balanced and FEMNIST. On EMNIST-Balanced under LGA, FedMAST achieves 93.01$\pm$1.12\% MTA and 0.22$\pm$0.18\% ASR over the 100-round run. On FEMNIST with method-aware \textsc{CovertLayers} attackers, FedMAST achieves 85.15$\pm$2.25\% MTA and 0.14$\pm$0.08\% ASR over the final 100 rounds.

\section{Discussion and Limitations}

The results support the multi-axis design across six attack families
and an adaptive variant, but this robustness comes at a cost. FedMAST incurs higher benign-rejection rates under stronger heterogeneity, with errors concentrated in a small set of partitions whose natural variation resembles attack evidence. Client-specific calibration may reduce this error concentration. FedMAST also requires stable pseudonymous identifiers for the partition-specific temporal state. While LGA and \textsc{CovertLayers} stress direction-aware and structure-aware evasion, identity-reset/Sybil behavior, and attackers directly optimizing against FedMAST's runtime spectral state are left for future evaluation. Broader validation across additional modalities, larger model architectures, stronger non-IID regimes, and longer-horizon adaptive attackers is needed to fully characterize generalization.

\section{Conclusion}

We presented FedMAST, a server-side defense that detects and contains stealth-constrained backdoor attacks by integrating structural, historical, and spectral evidence derived from the residual signatures these attacks leave behind. Signed temporal spectral drift captures persistent directional evidence, squeeze-pair coherence exposes compensating distortions across coupled features, and historical structure reveals updates that appear plausible within a round but deviate from trusted behavior over time. Across six stealth-constrained attacks and the method-aware \textsc{CovertLayers} attack, FedMAST limited malicious-update selection to 0--7.7\%, with zero malicious selection in five of the seven evaluated attacks while maintaining high main-task accuracy.

\FloatBarrier
\bibliographystyle{IEEEtran}
\bibliography{references}

@inproceedings{mcmahan2017communication,
  title={Communication-efficient learning of deep networks from decentralized data},
  author={McMahan, Brendan and Moore, Eider and Ramage, Daniel and Hampson, Seth and y Arcas, Blaise Aguera},
  booktitle={Artificial intelligence and statistics},
  pages={1273--1282},
  year={2017},
  organization={PMLR}
}

@inproceedings{guerraoui2018hidden,
  title={The hidden vulnerability of distributed learning in byzantium},
  author={Guerraoui, Rachid and Rouault, S{\'e}bastien and others},
  booktitle={International conference on machine learning},
  pages={3521--3530},
  year={2018},
  organization={PMLR}
}

@article{baruch2019little,
  title={A little is enough: Circumventing defenses for distributed learning},
  author={Baruch, Gilad and Baruch, Moran and Goldberg, Yoav},
  journal={Advances in Neural Information Processing Systems},
  volume={32},
  year={2019}
}

@inproceedings{bhagoji2019analyzing,
  title={Analyzing federated learning through an adversarial lens},
  author={Bhagoji, Arjun Nitin and Chakraborty, Supriyo and Mittal, Prateek and Calo, Seraphin},
  booktitle={International conference on machine learning},
  pages={634--643},
  year={2019},
  organization={PMLR}
}

@article{rieger2022deepsight,
  title={Deepsight: Mitigating backdoor attacks in federated learning through deep model inspection},
  author={Rieger, Phillip and Nguyen, Thien Duc and Miettinen, Markus and Sadeghi, Ahmad-Reza},
  journal={arXiv preprint arXiv:2201.00763},
  year={2022}
}

@inproceedings{zhang2022fldetector,
  title={Fldetector: Defending federated learning against model poisoning attacks via detecting malicious clients},
  author={Zhang, Zaixi and Cao, Xiaoyu and Jia, Jinyuan and Gong, Neil Zhenqiang},
  booktitle={Proceedings of the 28th ACM SIGKDD conference on knowledge discovery and data mining},
  pages={2545--2555},
  year={2022}
}

@inproceedings{huang2023multi,
  title={Multi-metrics adaptively identifies backdoors in federated learning},
  author={Huang, Siquan and Li, Yijiang and Chen, Chong and Shi, Leyu and Gao, Ying},
  booktitle={Proceedings of the IEEE/CVF International Conference on Computer Vision},
  pages={4652--4662},
  year={2023}
}

@article{wang2020attack,
  title={Attack of the tails: Yes, you really can backdoor federated learning},
  author={Wang, Hongyi and Sreenivasan, Kartik and Rajput, Shashank and Vishwakarma, Harit and Agarwal, Saurabh and Sohn, Jy-yong and Lee, Kangwook and Papailiopoulos, Dimitris},
  journal={Advances in neural information processing systems},
  volume={33},
  pages={16070--16084},
  year={2020}
}

@article{cao2020fltrust,
  title={Fltrust: Byzantine-robust federated learning via trust bootstrapping},
  author={Cao, Xiaoyu and Fang, Minghong and Liu, Jia and Gong, Neil Zhenqiang},
  journal={arXiv preprint arXiv:2012.13995},
  year={2020}
}

@inproceedings{ali2024adversarially,
  title={Adversarially guided stateful defense against backdoor attacks in federated deep learning},
  author={Ali, Hassan and Nepal, Surya and Kanhere, Salil S and Jha, Sanjay},
  booktitle={2024 Annual Computer Security Applications Conference (ACSAC)},
  pages={794--809},
  year={2024},
  organization={IEEE}
}

@article{wan2026mars,
  title={Mars: A malignity-aware backdoor defense in federated learning},
  author={Wan, Wei and Yuxuan, Ning and Huang, Zhicong and Hong, Cheng and Hu, Shengshan and Zhou, Ziqi and Zhang, Yechao and Zhu, Tianqing and Zhou, Wanlei and Zhang, Leo Yu},
  journal={Advances in Neural Information Processing Systems},
  volume={38},
  pages={154985--155017},
  year={2026}
}

@article{hsu2019measuring,
  title={Measuring the effects of non-identical data distribution for federated visual classification},
  author={Hsu, Tzu-Ming Harry and Qi, Hang and Brown, Matthew},
  journal={arXiv preprint arXiv:1909.06335},
  year={2019}
}

@inproceedings{xie2021crfl,
  title={Crfl: Certifiably robust federated learning against backdoor attacks},
  author={Xie, Chulin and Chen, Minghao and Chen, Pin-Yu and Li, Bo},
  booktitle={International conference on machine learning},
  pages={11372--11382},
  year={2021},
  organization={PMLR}
}

@article{fereidooni2023freqfed,
  title={Freqfed: A frequency analysis-based approach for mitigating poisoning attacks in federated learning},
  author={Fereidooni, Hossein and Pegoraro, Alessandro and Rieger, Phillip and Dmitrienko, Alexandra and Sadeghi, Ahmad-Reza},
  journal={arXiv preprint arXiv:2312.04432},
  year={2023}
}

@article{abacha2026fedsurrogate,
  title={FedSurrogate: Backdoor Defense in Federated Learning via Layer Criticality and Surrogate Replacement},
  author={Abacha, Fatima Z and Teo, Sin G and Wu, Yuanxiang and Cordeiro, Lucas C and Mustafa, Mustafa A},
  journal={arXiv preprint arXiv:2605.11122},
  year={2026}
}

@inproceedings{bagdasaryan2020backdoor,
  title={How to backdoor federated learning},
  author={Bagdasaryan, Eugene and Veit, Andreas and Hua, Yiqing and Estrin, Deborah and Shmatikov, Vitaly},
  booktitle={International conference on artificial intelligence and statistics},
  pages={2938--2948},
  year={2020},
  organization={PMLR}
}

@inproceedings{zhang2022neurotoxin,
  title={Neurotoxin: Durable backdoors in federated learning},
  author={Zhang, Zhengming and Panda, Ashwinee and Song, Linyue and Yang, Yaoqing and Mahoney, Michael and Mittal, Prateek and Kannan, Ramchandran and Gonzalez, Joseph},
  booktitle={International conference on machine learning},
  pages={26429--26446},
  year={2022},
  organization={PMLR}
}

@inproceedings{li20233dfed,
  title={3dfed: Adaptive and extensible framework for covert backdoor attack in federated learning},
  author={Li, Haoyang and Ye, Qingqing and Hu, Haibo and Li, Jin and Wang, Leixia and Fang, Chengfang and Shi, Jie},
  booktitle={2023 IEEE symposium on security and privacy (SP)},
  pages={1893--1907},
  year={2023},
  organization={IEEE}
}

@inproceedings{zhuang2024backdoor,
  title={Backdoor federated learning by poisoning backdoor-critical layers},
  author={Zhuang, Haomin and Yu, Mingxian and Wang, Hao and Hua, Yang and Li, Jian and Yuan, Xu},
  booktitle={International Conference on Learning Representations},
  volume={2024},
  pages={40241--40266},
  year={2024}
}

@inproceedings{xie2019dba,
  title={Dba: Distributed backdoor attacks against federated learning},
  author={Xie, Chulin and Huang, Keli and Chen, Pin-Yu and Li, Bo},
  booktitle={International conference on learning representations},
  year={2019}
}

@inproceedings{yang2025stealthy,
  title={Stealthy Backdoor Attack in Federated Learning via Adaptive Layer-wise Gradient Alignment},
  author={Yang, Qingqian and Yan, Peishen and Wu, Xiaoyu and Zhang, Jiaru and Song, Tao and Hua, Yang and Wang, Hao and Wang, Liangliang and Guan, Haibing},
  booktitle={Proceedings of the IEEE/CVF International Conference on Computer Vision},
  pages={29163--29172},
  year={2025}
}

@inproceedings{xu2025detecting,
  title={Detecting backdoor attacks in federated learning via direction alignment inspection},
  author={Xu, Jiahao and Zhang, Zikai and Hu, Rui},
  booktitle={Proceedings of the Computer Vision and Pattern Recognition Conference},
  pages={20654--20664},
  year={2025}
}

@inproceedings{nguyen2022flame,
  title={$\{$FLAME$\}$: Taming backdoors in federated learning},
  author={Nguyen, Thien Duc and Rieger, Phillip and Chen, Huili and Yalame, Hossein and M{\"o}llering, Helen and Fereidooni, Hossein and Marchal, Samuel and Miettinen, Markus and Mirhoseini, Azalia and Zeitouni, Shaza and others},
  booktitle={31st USENIX security symposium (USENIX Security 22)},
  pages={1415--1432},
  year={2022}
}

@article{blanchard2017machine,
  title={Machine learning with adversaries: Byzantine tolerant gradient descent},
  author={Blanchard, Peva and El Mhamdi, El Mahdi and Guerraoui, Rachid and Stainer, Julien},
  journal={Advances in neural information processing systems},
  volume={30},
  year={2017}
}

@inproceedings{yin2018byzantine,
  title={Byzantine-robust distributed learning: Towards optimal statistical rates},
  author={Yin, Dong and Chen, Yudong and Kannan, Ramchandran and Bartlett, Peter},
  booktitle={International conference on machine learning},
  pages={5650--5659},
  year={2018},
  organization={Pmlr}
}

@inproceedings{zhang2026heterogeneity,
  title={Heterogeneity-Oblivious Robust Federated Learning},
  author={Zhang, Weiyao and Li, Jinyang and Song, Qi and Wang, Miao and Lin, Chungang and Luo, Haitong and Meng, Xuying and Zhang, Yujun},
  booktitle={IEEE INFOCOM 2026-IEEE Conference on Computer Communications},
  pages={1--10},
  year={2026},
  organization={IEEE}
}

@misc{fedmast_artifact,
  author    = {Srinivasan Subramanian and Md Abdullah Al Hafiz Khan and Kazi Aminul Islam},
  title     = {{FedMAST}: Backdoor Detection and Containment in Federated Learning},
  year      = {2026},
  publisher = {Zenodo},
  doi       = {10.5281/zenodo.22885416},
  url       = {https://doi.org/10.5281/zenodo.22885416}
}

\appendix
\subsection{Update Representation and Feature Bank}
FedMAST computes features from trainable model parameters. For the Tiny ResNet-18 used on CIFAR-10, the model is divided into six stages,
\[
\{\mathrm{stem},\mathrm{layer1},\mathrm{layer2},
  \mathrm{layer3},\mathrm{layer4},\mathrm{head}\}.
\]
Let $\Delta_i$ denote the flattened update from client $i$ and $\Delta_i^{(s)}$ its update for stage $s$. For the clients sampled in round $t$,
\[
\bar{\Delta}_t =
\frac{1}{|S_t|}\sum_{j\in S_t}\Delta_j,
\qquad
\bar{\Delta}_{-i,t} =
\frac{1}{|S_t|-1}
\sum_{\substack{j\in S_t\\j\neq i}}\Delta_j.
\]

Table~\ref{tab:app-feature-defs} summarizes the 22 structural features.
The historical-baseline quantities are based on the historical statistics available to FedMAST. In our primary experiments, these statistics are initialized from a previous 50-round attack-free history. When such a history is not available, FedMAST builds the corresponding baselines from accepted updates during training. For a reshaped weight tensor to $[\mathrm{out},-1]$ with singular values $\sigma_1,\ldots,\sigma_r$, the spectral quantities are
\[
R_{\mathrm{SV}}
=
\frac{\sigma_1}{\sum_j\sigma_j+10^{-12}},
\qquad
H_{\mathrm{SV}}
=
-\sum_j p_j\log p_j,
\quad
\]
\[
p_j=
\frac{\sigma_j}{\sum_k\sigma_k+10^{-12}}.
\]
When a stage contains several weight tensors, the corresponding spectral
quantity is combined using the Frobenius-norm weighting.

The signed spectral-drift path uses six additional quantities, such as
top-singular-value ratio and spectral entropy from layers 2, 3, and 4.
These quantities are kept separate from the structural feature bank so
that their direction is preserved. The per-appearance spectral evidence is as follows.
\[
s_{i,t}
=
-\operatorname{mean}(z_{\mathrm{topSV}})
+
\operatorname{mean}(z_{\mathrm{entropy}}).
\]
At least two valid layer values are required for each term. FedMAST uses the following four predefined squeeze pairs.
\[
\begin{aligned}
&(\text{baseline displacement},\ \text{layer-4 energy}),\\
&(\text{round-mean alignment},\ \text{head spectral entropy}),\\
&(\text{head-energy ratio},\ \text{head top-SV ratio}),\\
&(\text{classifier/backbone ratio},\ \text{backbone kurtosis}).
\end{aligned}
\]

\subsection{FedMAST Hyperparameters}
Table~\ref{tab:app-hparams} lists the FedMAST hyperparameters. These values are configurable and are not requirements of the scoring procedure.

\begin{table}[!htbp]
\centering
\caption{FedMAST implementation defaults and decision constants.}
\label{tab:app-hparams}
\scriptsize
\setlength{\tabcolsep}{3.0pt}
\renewcommand{\arraystretch}{1.03}
\begin{tabular}{@{}p{0.61\linewidth}p{0.28\linewidth}@{}}
\toprule
\textbf{Setting} & \textbf{Value} \\
\midrule
MAD multiplier $k$ & 3 \\
Hard-axis consensus $c$ & 2 of 3 \\
Target minimum accepted clients $n_{\min}$ & 5 \\
Hard safety floor $n_{\mathrm{floor}}$ & 3 \\
Strong hard rejection & 3 of 3, one axis $>2\times$ threshold \\
Warmup without stored history & 2 rounds \\
Rolling accepted-history buffer & 20 rounds \\
Anchor disable threshold $\theta_{\mathrm{cat}}$ & 100 \\
\midrule
Trajectory window & 15 appearances \\
Trajectory EMA new-value weight & 0.3 \\
Momentum decay & 0.7 \\
Momentum activation & 5 appearances \\
\midrule
Spectral EMA decay $\alpha$ & 0.7 \\
Spectral activation $n_a$ & 5 appearances \\
Spectral percentile $q$ & 85 \\
Minimum spectral threshold & 0.5 \\
\midrule
Minimum clients for cluster split & 5 \\
Minimum available history features & 5 \\
Minimum clients per cluster & 2 \\
Cluster distance-ratio gate & 1.5 \\
\midrule
Accepted-aggregate EMA decay & 0.8 \\
Aggregate-drift threshold & 0.3 \\
Suspicious-round containment & Coordinate median \\
Suspicious-round trimmed-mean fraction & 0.2 \\
\bottomrule
\end{tabular}
\end{table}

The three hard axes are
$D_{\mathrm{round}}$, $D_{\mathrm{squeeze}}$, and $D_{\mathrm{hist}}$.
A client is hard-flagged when at least two of the three axes exceed their
independent robust thresholds.

$D_{\mathrm{spectral}}$ is evaluated separately. Once active, a spectral
flag is not eligible for the minimum-acceptance rescue.

$D_{\mathrm{traj}}$, $D_{\mathrm{momentum}}$, and $D_{\mathrm{anchor}}$
are soft evidence. They contribute to ranking, historical consistency
checks, and re-selection, but do not independently cause ordinary hard
rejection.

\subsection{Historical State}

The primary CIFAR-10 experiments use a frozen 50-round attack-free FedAvg
history to initialize the historical baselines. The history contains the
robust feature statistics and aggregate structural state required by the
historical scoring paths. FedMAST does not require a supplied history. When no frozen history is available, a two-round warmup is used and the same baseline statistics are built from the rolling buffer of accepted low-risk updates. Only reliable rounds update this rolling history. A round is quarantined
when FedMAST marks it suspicious or detects elevated spectral risk.

\begin{table*}[!htbp]
\centering
\caption{Structural features used by FedMAST.}
\label{tab:app-feature-defs}
\scriptsize
\setlength{\tabcolsep}{2.3pt}
\renewcommand{\arraystretch}{1.02}
\begin{tabular}{@{}p{0.15\textwidth}p{0.24\textwidth}p{0.56\textwidth}@{}}
\toprule
\textbf{Family} & \textbf{Feature} & \textbf{Quantity or role} \\
\midrule

Magnitude
& Update norm
& $\|\Delta_i\|_2$ \\

& Distance to round mean
& $\|\Delta_i-\bar{\Delta}_t\|_2$ \\

& Distance to baseline update
& Displacement from the historical baseline \\

& Cosine to round mean
& Cosine similarity between $\Delta_i$ and $\bar{\Delta}_t$ \\
\midrule

Alignment
& Leave-one-out cosine
& Cosine similarity between $\Delta_i$ and $\bar{\Delta}_{-i,t}$ \\

& Cosine to baseline update
& Directional agreement with the historical baseline \\

& Stage-norm cosine
& Cosine between the six stage norms of $\Delta_i$ and the corresponding
round-mean stage norms \\

& Head-sign agreement
& Fraction of non-negligible classifier-head entries with matching signs \\
\midrule

Layer energy
& Head norm
& $\|\Delta_i^{(\mathrm{head})}\|_2$ \\

& Layer-4 norm
& $\|\Delta_i^{(\mathrm{layer4})}\|_2$ \\

& Head/total ratio
& $\|\Delta_i^{(\mathrm{head})}\|_2/(\|\Delta_i\|_2+10^{-12})$ \\

& Classifier/backbone ratio
& Head norm divided by the combined backbone-stage norms \\

& Head--backbone product
& Head-energy ratio combined with maximum backbone kurtosis \\
\midrule

Spectral/shape
& Head spectral entropy
& Entropy of the singular-value distribution of the classifier update \\

& Head top-SV ratio
& Largest singular value divided by the sum of singular values \\

& Layer-4 skewness
& Third standardized moment of the flattened layer-4 update \\

& Layer-3 kurtosis
& Excess kurtosis of the flattened layer-3 update \\

& Maximum backbone kurtosis
& Maximum excess kurtosis over backbone stages \\
\midrule

Cross-layer
& Layer-4 cosine
& Layer-4 cosine similarity to the round mean \\

& Head leave-one-out cosine
& Classifier-head cosine similarity to the leave-one-out mean \\

& Class-update entropy
& Entropy of classifier-row update magnitudes \\

& Layer-4 $\ell_\infty/\ell_2$
& $\|\Delta_i^{(\mathrm{layer4})}\|_\infty/
(\|\Delta_i^{(\mathrm{layer4})}\|_2+10^{-12})$ \\
\bottomrule
\end{tabular}
\end{table*}

\section{Experimental Configuration}
\label{app:configuration}
\subsection{Initialization, Trigger, and Evaluation}
All primary CIFAR-10 runs use the same Tiny ResNet-18 global-model
initialization and the same client partitions. The initialization is fixed
across defenses so that comparisons begin from the same global model. Main-task accuracy is evaluated on the standard CIFAR-10 test split after every communication round.

Unless an attack defines its own trigger construction, the common backdoor
trigger is an $8\times8$ orange patch placed in the bottom-right corner of
the image with the target label 2. The standard poisoned-data path replaces
20 examples per poisoned minibatch. ASR excludes test examples whose true
class is already the target class.

DBA uses a different trigger construction. Four persistent attackers each
train on a separate six-pixel local trigger. The global DBA trigger is the
union of all four local patterns, giving a 24-pixel trigger that is not
observed by any one attacker during local training. DBA ASR is evaluated
with this assembled global trigger.

\subsection{Primary Attack Configurations}

Table~\ref{tab:app-attack-config} records the main attack-specific settings.
The reported start round is configured as attack-start. DBA begins the attack only after the simulator has identified all four fixed attacker partitions.

\begin{table}[!htbp]
\centering
\caption{Baseline-defense configurations.}
\label{tab:app-baselines}
\scriptsize
\setlength{\tabcolsep}{2.6pt}
\renewcommand{\arraystretch}{1.04}
\begin{tabular}{@{}p{0.25\linewidth}p{0.66\linewidth}@{}}
\toprule
\textbf{Defense} & \textbf{Configuration} \\
\midrule

Trimmed Mean
& Coordinate-wise trimming with fraction $0.2$ from each tail. \\

MultiKrum
& Byzantine budget $f$ equals the configured malicious-client count for
the run. Krum uses $n-f-2$ neighbors and MultiKrum retains the five
lowest-score updates. \\

FLAME
& Cosine-distance HDBSCAN with minimum cluster size
$\lfloor n/2\rfloor+1$ and \texttt{min\_samples}=1.
The clipping bound is the median norm over all client updates.
Selected updates are clipped and averaged, followed by per-coordinate
Gaussian noise with standard deviation $0.001$ times the clipping bound. \\

AlignIns
& Significant-coordinate fraction $0.3$,
$\lambda_s=1.0$, and $\lambda_c=1.0$.
Clients are filtered using the MPSA and TDA standardized scores and the
retained updates are clipped using their median update norm. \\
\bottomrule
\end{tabular}
\end{table}

\subsection{Baseline Configurations}

All baseline defenses use the same sampled clients, partitions, attack
configuration, and global-model initialization as the corresponding
FedMAST experiment. Table~\ref{tab:app-baselines} lists the aggregation
settings used.

\section{Supplementary Analysis}
\subsection{Axis Coverage}
Table~\ref{tab:axis-coverage} measures each axis before consensus. LGA
mainly exceeds $D_{\mathrm{spec}}$, while the other attacks expose
multiple structural traces. This variation motivates multi-axis evidence
rather than reliance on a single anomaly signal.

\begin{table}[!htbp]
\centering
\caption{Axis-wise malicious coverage before consensus (\%).}
\label{tab:axis-coverage}
\scriptsize
\setlength{\tabcolsep}{3.4pt}
\renewcommand{\arraystretch}{1.08}
\begin{tabular}{@{}lcccc@{}}
\toprule
Attack & $D_{\mathrm{round}}$ & $D_{\mathrm{squeeze}}$ & $D_{\mathrm{hist}}$ & $D_{\mathrm{spec}}$ \\
\midrule
Constrain \& Scale & 86.5 & 90.5 & 85.5 & 64.5 \\
Neurotoxin & 100.0 & 99.5 & 100.0 & 74.5 \\
BC-Layers & 100.0 & 100.0 & 100.0 & 28.0 \\
LGA & 1.7 & 1.1 & 1.8 & 89.0 \\
DBA & 95.6 & 77.9 & 95.4 & 34.6 \\
3DFed & 100.0 & 100.0 & 100.0 & 0.0 \\
\textsc{CovertLayers} & 100.0 & 99.8 & 100.0 & 51.5 \\
\bottomrule
\multicolumn{5}{l}{\footnotesize Values report the percentage of malicious updates that} \\
\multicolumn{5}{l}{\footnotesize independently exceed each axis threshold.}
\end{tabular}
\end{table}

\begin{table*}[!htbp]
\centering
\caption{Primary CIFAR-10 attack configurations.}
\label{tab:app-attack-config}
\scriptsize
\setlength{\tabcolsep}{2.8pt}
\renewcommand{\arraystretch}{1.05}
\begin{tabular}{@{}p{0.13\textwidth}ccp{0.65\textwidth}@{}}
\toprule
\textbf{Attack} &
\textbf{Mal./round} &
\textbf{Start} &
\textbf{Configuration} \\
\midrule

C\&S
& 1 & 1
& 3 attack epochs, LR $0.01$, scale $1.0$, proximity weight $0.5$,
poison ratio $0.5$ \\

Neurotoxin
& 1 & 1
& Heavy-hitter mask ratio $0.01$, update scale $1.0$ \\

BC-Layers
& 1 & 1
& $\tau=0.95$, $\lambda=1.0$, 5 proxy models,
2 benign epochs, 2 malicious epochs, LR $0.1$ \\

LGA
& 3 & 20
& 12 attack epochs, LR $0.05$, layer-wise alignment parameter $\tau=2.0$ \\

DBA
& 4 & 1
& Four persistent attackers, 6 attack epochs, LR $0.05$,
poison ratio $5/64$, update scale $1.0$, server $\eta=1.0$ \\

3DFed
& 3 & 30
& 2 backdoor models and 1 decoy,
$\beta=0.3$, $\gamma=1.0$, $\kappa=10^5$,
20 noise steps and 20 decoy steps with LR $0.01$ \\

\textsc{CovertLayers}
& 1 & 30
& BC-Layers base attack with $\tau=0.95$, $\lambda=0.7$, 5 proxies,
norm-match weight $0.8$, non-critical-layer smoothing noise $0.03$,
direction blend $0.15$, minimum BSR ratio $0.5$ \\
\bottomrule
\end{tabular}
\end{table*}

\subsection{Round-Level Detection Stability}

We reconstruct the client-level decision in every attacker-active round to capture whether malicious selections are distributed across training or concentrated in a small number of rounds.

A \emph{perfect round} rejects every malicious update while rejecting no
benign update. A \emph{zero-catch round} rejects none of the malicious
updates in that round.

\begin{table}[!htbp]
\centering
\caption{Round-level detection behavior in the primary CIFAR-10 traces.
Perfect and zero-catch rates use attacker-active rounds. FPR uses benign
appearances over the full run.}
\label{tab:app-round-stability}
\scriptsize
\setlength{\tabcolsep}{2.8pt}
\renewcommand{\arraystretch}{1.05}
\begin{tabular}{@{}lrrrr@{}}
\toprule
Attack & Recall & FPR & Perfect & Zero-catch \\
       & (\%) & (\%) & rounds (\%) & rounds (\%) \\
\midrule
C\&S       & 96.50  & 16.28 & 21.00 & 3.50 \\
Neurotoxin & 100.00 & 10.44 & 32.50 & 0.00 \\
BC-Layers  & 100.00 & 15.44 & 18.00 & 0.00 \\
LGA        & 92.27  & 23.75 & 9.39  & 2.76 \\
DBA        & 100.00 & 4.23  & 98.86 & 0.00 \\
3DFed      & 100.00 & 9.82  & 58.48 & 0.00 \\
\textsc{CovertLayers}
            & 100.00 & 14.65 & 21.05 & 0.00 \\
\bottomrule
\end{tabular}
\end{table}

DBA and 3DFed do not have zero-catch rounds after their attackers become
active. LGA's malicious update recall is 92.27\%, but only 9.39\% of its attacker-active rounds reject all malicious updates without also rejecting a benign update.

\subsection{Containment Statistics}

\begin{table}[!htbp]
\centering
\caption{Aggregation policy during attacker-active rounds.}
\label{tab:app-round-policy}
\scriptsize
\setlength{\tabcolsep}{3.4pt}
\renewcommand{\arraystretch}{1.05}
\begin{tabular}{@{}lrr@{}}
\toprule
Attack & FedAvg (\%) & Median (\%) \\
\midrule
C\&S       & 19.00 & 81.00 \\
Neurotoxin & 22.50 & 77.50 \\
BC-Layers  & 43.50 & 56.50 \\
LGA        & 1.66  & 98.34 \\
DBA        & 0.00  & 100.00 \\
3DFed      & 43.86 & 56.14 \\
\textsc{CovertLayers}
            & 32.16 & 67.84 \\
\bottomrule
\end{tabular}
\end{table}

FedMAST uses FedAvg over the accepted set when the round is considered
reliable. When the round is suspicious, the primary experiments instead
use coordinate-median aggregation over the accepted updates. Table~\ref{tab:app-round-policy} reports the resulting policy frequencies during attacker-active rounds. These results show that containment is an active part of defense and it is clearly visible on LGA and DBA.

\subsection{Predefined Squeeze Pairs}

\begin{table}[!htbp]
\centering
\caption{Percentile rank of the strongest predefined squeeze pair among
all 231 structural-feature pairs in each retained trace.}
\label{tab:app-squeeze-percentile}
\scriptsize
\setlength{\tabcolsep}{2.3pt}
\renewcommand{\arraystretch}{1.05}
\begin{tabular}{@{}l p{0.48\linewidth} r@{}}
\toprule
Attack & Strongest predefined pair & Pct. \\
\midrule
C\&S
& Baseline displacement + layer-4 energy & 98.3 \\

Neurotoxin
& Baseline displacement + layer-4 energy & 87.0 \\

LGA
& Head energy + head spectral rank & 99.6 \\

DBA
& Baseline displacement + layer-4 energy & 92.2 \\

3DFed
& Cross-layer ratio + distributional shape & 90.5 \\

\textsc{CovertLayers}
& Baseline displacement + layer-4 energy & 84.8 \\
\bottomrule
\end{tabular}
\end{table}

To examine whether the four predefined squeeze pairs capture useful joint
structure, we compare them post hoc with all possible pairs from the structural feature bank. Each pair is ranked by its pooled-covariance two-dimensional separation between benign and malicious updates. This diagnostic is used only to compare candidate feature pairs. 

Four of the six retained traces place their strongest predefined pair
above the 90th percentile of all candidate pairs. The strongest pair is
also attack-dependent, which supports using several fixed structural
pairs instead of relying on one pair for all attack mechanisms. We do not use the ratio between two-dimensional separation and the best one-dimensional separation as evidence for the squeeze mechanism, because that ratio is not an informative test of complementarity for this distance construction.

\end{document}